\documentclass[sigconf]{acmart}

\usepackage{booktabs} % For formal tables
\usepackage{epsfig}
\usepackage{graphicx}
\usepackage{amsmath}
\usepackage{amssymb}
\usepackage{multirow}
\usepackage{bbm}
\usepackage{float}
\usepackage{balance}
\setcopyright{acmlicensed}
\begin{document}
\title{Modeling Image Virality \\with Pairwise Spatial Transformer Networks}

\author{Abhimanyu Dubey}
\affiliation{%
  \institution{Massachusetts Institute of Technology}
}
\email{dubeya@mit.edu}

\author{Sumeet Agarwal}
\affiliation{%
  \institution{Indian Institute of Technology Delhi}
}
\email{sumeet@iitd.ac.in}
% \renewcommand{\shorttitle}{Modeling Virality with Pairwise STNs}
% \renewcommand{\shortauthors}{Dubey and Agarwal}

\begin{abstract}
The study of virality and information diffusion is a topic gaining traction rapidly in the computational social sciences. Computer vision and social network analysis research have also focused on understanding the impact of content and information diffusion in making content viral, with prior approaches not performing significantly well as other traditional classification tasks. In this paper, we present a novel pairwise reformulation of the virality prediction problem as an attribute prediction task and develop a novel algorithm to model image virality on online media using a pairwise neural network. Our model provides significant insights into the features that are responsible for promoting virality and surpasses the existing state-of-the-art by a 12\% average improvement in prediction. We also investigate the effect of external category supervision on relative attribute prediction and observe an increase in prediction accuracy for the same across several attribute learning datasets.
\end{abstract}
\copyrightyear{2017} 
\acmYear{2017} 
\setcopyright{acmlicensed}
\acmConference{MM'17}{}{October 23--27, 2017, Mountain View, CA, USA.}
 \acmPrice{15.00}
 \acmDOI{https://doi.org/10.1145/3123266.3123333}
 \acmISBN{ISBN 978-1-4503-4906-2/17/10}


\maketitle
\clubpenalty=10000 
\widowpenalty = 10000 
\section{Introduction}
Online advertising has changed form persistently since the dawn of easily available Internet connectivity, beginning from blunt header and footer advertisements to carefully curated and unidentifiable advertisements fused closely with content. A call-and-response mechanism has started to emerge in the propagation of content on the Internet - content aggregator websites such as BuzzFeed \cite{buzzfeed}, Gawker \cite{gawker} and Funny or Die \cite{funnyordie} immediately capture  the \textit{viral} content emerging on websites such as Reddit \cite{reddit} and a subsequent marketing cascade follows that builds on the \textit{virality} of content.\\

Apart from online marketing, the impact of several other domains of active Internet participation depends on content virality. The reach of professionals, organizations, social causes and non-profits spitballs exponentially once viral content is associated with the same. Hence, as described previously in Deza and Parikh's \cite{deza2015} novel introductory study of image virality, content virality has been studied extensively in the domain of marketing research \cite{mr1,mr2,mr3,mr4,mr5}.\\

It is important to note the subtle difference between viral content and content that is otherwise generally popular on the Internet. Earlier work by Khosla et. al. \cite{khosla2014} aims at understanding the visual cues that make an image \textit{popular} on the Internet. However, it is understandable that content that is popular is not necessarily viral, and dissemination networks are significantly different for the two classes of content. Deza and Parikh's work is an important stepping stone to understanding the nature of content virality, and Lakkaraju et. al. \cite{lakkaraju2013} describe the temporal relationships of image virality in mode detail, along with several other streams of research \cite{jain2014,goel2015} that discuss the nature of the underlying structure of diffusion present in viral content. This posits the obvious question of the relative importance of the content matter of a viral image, and if it is content alone that can govern the extent of virality an image gains online. Deza and Parikh perform an extensive study of the same, using handcrafted computer vision techniques - identifying that it is possible, with a certain degree of accuracy, to predict the virality of an image based on the image content alone.\\

Across the multimedia and Internet media research communities, we have seen a surge in the usage of deep learning for end-to-end learning of complicated  tasks. However, all these tasks have utilized computational models that concrete ontological categories, such as image classification \cite{krizhevsky2012}, region proposals and semantic segmentation \cite{girshick2014}, image captioning \cite{lin2014} and visual question answering \cite{antol2015}. Extending traditional computational models to tasks that involve amorphous, abstract ontologies - such as image virality and popularity (that have large inter-user disagreements in label data), we require a rethinking of the problem approach, as done by prior work in image memorability \cite{isola2011,khosla2012,khosla22012}, interestingness \cite{turakhia2013,gygli2013} and urban perception \cite{dubey2016deep,naik2014}.\\

One of our most important contributions through this work is to demonstrate the effectiveness of a pairwise attribute approach in tasks that involve an amorphous ontology (for example, image popularity, beauty, memorability). We demonstrate a hierarchical novel approach, that utilizes transfer learning from traditional concrete domains, and provide substantial empirical improvements in prediction. Our approach is generalizable across abstract categories, and also outperforms prior work in attribute learning as well.
\section{Related Work}
\subsection{Predicting Content Popularity}
A variety of different disciplines from computational social science to computer vision have approached the problem of predicting content popularity on the Internet. In the domains of machine learning and computer vision, most of the work currently present has been in the domain of Twitter or video - predicting the popularity of tweets \cite{petrovic2011,hong2011} and videos \cite{pinto2013,shamma2011,nwana2013}. However, there is a growing body of work on predicting image popularity and virality \cite{deza2015,khosla2014}, with the work of Khosla et al \cite{khosla2014}, which focuses on introducing the problem of predicting image popularity, and then presenting a straightforward formulation with image features applied to a classifier. Deza and Parikh \cite{deza2015} introduce the different problem of predicting image virality with a novel metric for identifying virality, and demonstrate the effectiveness of several techniques for solving the problem using deep image attributes. Our work uses the datasets introduced by these studies and surpasses their proposed solutions by a large average improvement.\\

Social Media research has also looked at predicting content virality - focusing on the underlying structure of information dissemination in a social network and the cascading effect that aggregator websites have. They explain the nature of virality as a network phenomenon, and not entirely attributable to the inherent content itself \cite{bakshy2012,jain2014,wihbey2014}. Jain et al. \cite{jain2014} use an intricate system to identify viral videos in real-time. They accomplish this by analyzing a metric based on the number of re-shares a particular video has on Twitter and select the top-500 videos according to their metric. Goel et al. \cite{goel2015} model the spread of viral ideas as a diffusion process similar to the spread of an infectious disease \cite{anderson1991,coleman1957,bass1969}, and introduce a formal metric for `structural virality', which accounts for the spread of information through both large broadcasts (i.e. sources with a large audience), as well as multiple re-sharing with smaller individual audiences, and evaluate the metric on an extremely large dataset of a billion events on Twitter. Their findings are contradictory to \cite{jain2014}, since they posit that structural virality is dependent primarily on the size of the largest individual broadcast (person with the most number of followers) and is low throughout.
\subsection{Attribute Learning}
Beginning with the seminal work of Parikh and Grauman \cite{parikh2011} in 2011, predicting and understanding relative attributes in images is becoming an area of active research. In their paper, they introduce the importance of identifying `relative' attributes in images, and propose a solution based on shallow image features and the RankSVM \cite{joachims2002} formulation propsed by Joachims in 2002. There has been significant subsquent improvements to the original formulation, such as the work of Souri et al \cite{souri2015}, who extend the original relative attribute learning formulation using a ranking network introduced by Cao et al \cite{cao2007learning}.\\

Singh et al \cite{singh2016end} propose a novel algorithm for relative attribute prediction using spatial transformer networks \cite{jaderberg2015spatial} in conjunction with a siamese architecture \cite{chopra2005} to learn relative attribute prediction using the ranking loss proposed by \cite{cao2007learning}. They achieve the current state-of-the-art in realtive attribute prediction, surpassing the competing approaches of Xiao and Lee \cite{xiao2015discovering} which attempt to learn spatial extents of attributes using ensemble image representations. Our work extends research in this domain by demonstrating the applicability of these pairwise techniques in prediction tasks which are otherwise modelled as classification and regression, and the introduction of transfer learning into the domain of solving relative attribute prediction.\\

Our motivation, through this paper, is to provide a pipeline for understanding the nature of content virality. We understand that virality is a function of both content and the nature of the network, however, we tackle the problem of understanding the information - specifically, visual cues that can discriminate between viral and non-viral content. Our technical contribution is primarily to provide a general pairwise comparative deep architecture to learn a classification over the image set based on the strength of a visual attribute. We also aim to account for the disparity in the network topology present in the spread of viral information by creating a metric and dataset that balances the network effect. In the further sections, we present a more formal definition of the problem, and provide experimental results as well.
\section{Approach}
In \cite{deza2015}, the authors introduce an image dataset for predicting image virality. This dataset has images along with a metric encapsulating the virality of the image, calculated by taking into account the number of reshares over a short amount of time, which is characteristic to viral images. We also experiment with the image dataset introduced by \cite{khosla2014} to predict image popularity, details of which are explained in Section \ref{datasets}. We found that approaching these problems as regression tasks, where we aim to predict the virality/popularity metric for each image, was diffult to learn with considerable accuracy. This occured primarily because the labels themselves were varying in scale and were better understood in a relative sense.
\subsection{Problem Formulation}
Given the relative nature of the labels, we formulate the problem of virality prediction as a pairwise relative attribute prediction task. More specifically, we formulate a 2AFC (2 Alternative Forced Choice) problem, where the taask is to predict the more viral or more popular among a pair of images. We find this approach to be far more effective at predicting virality and popularity, since it eliminates variations in scale across samples. Our inference task can be explicitly stated as follows - for a pair of images $(I_1,I_2) $ in our set of total images $\mathbf{I}$, and a relative label $y \in \{l,r\}$, we take a set of input triplets $(I_1,I_2,y)$ during training time as input and predict the image with the stronger attribute (one of left or right) during test.
\subsection{Architecture}
Jaderberg et al \cite{jaderberg2015spatial} in 2015 introduce a powerful class of neural networks known as spatial transformer networks (STNs), which are aimed at learning an affine transformation of the input from the label provided. These architectures have been applied with success for virality prediction by \cite{singh2016end}. We aim to select regions of the image corresponding to areas of interest for virality prediction. We can formulate the task of STNs as learning an affine transformation $\mathbf A_\theta$ of the form
\begin{equation}
\mathbf A_\theta = \begin{bmatrix} s & 0 & t_x \\ 0 & s & t_y \end{bmatrix}
\end{equation}
The parameters $s, t_x, t_y$ can adjust the transformation by adjusting the scale and translation across $x$ and $y$ axes respectively. As described in detail by \cite{jaderberg2015spatial}, these parameters are learnt via backpropagation.\\

\textbf{Siamese Spatial Transformer Networks:} The basic architecture is described as follows - the CNN \textbf{ViralNet} takes in inputs $(\mathbf I_1, \mathbf I_2)$ and label $y$. $N$ is formed of two subnetworks $N_1,N_2$, which have identical weights and produce scalar output $v(\mathbf I_1)$ and $v(\mathbf I_2)$ each. Each of these subnetworks $N_i$ is composed of two subnetworks each - a spatial transformer network $S$ and a ranking network $R$.\\

\begin{figure*}[t]
\centering
\includegraphics[width=1\textwidth]{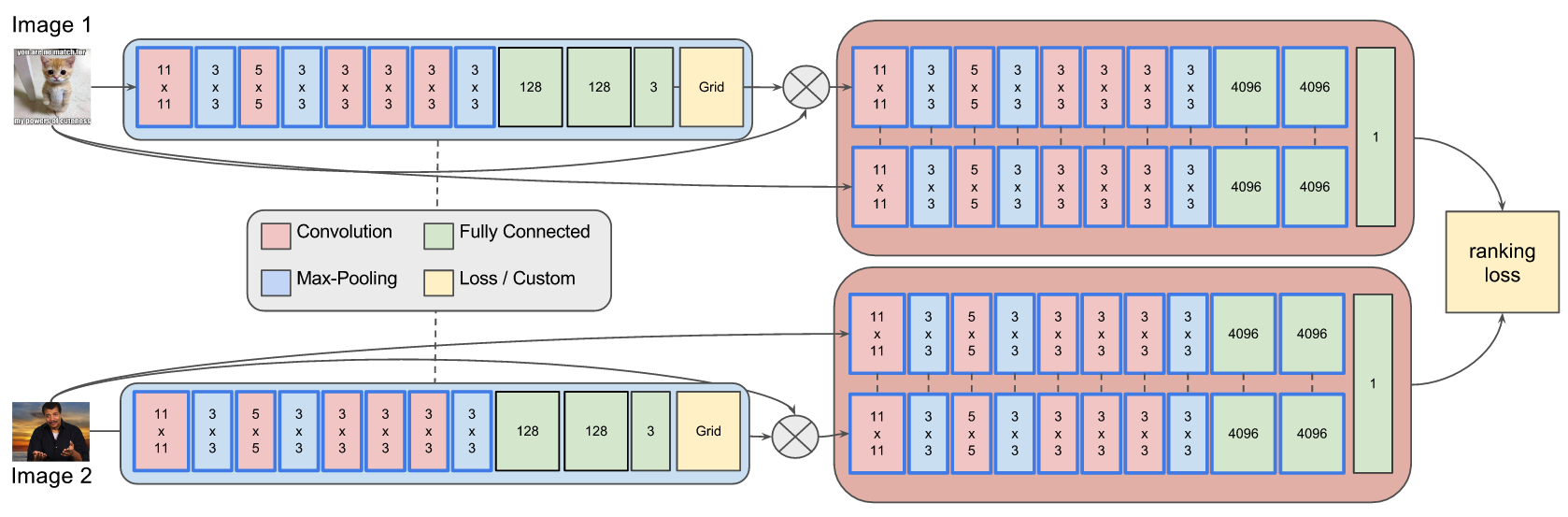}
\vspace{-10pt}
\caption{Full architecture for \textbf{ViralNet}. The first network, shaded in blue, represents the Spatial Transformer Network, whereas the consequent network, shaded in red, represents the Ranking Network. Note that both Spatial Transformer Network $S$ and Ranking Network $R$ share weights for both input images. Layers outlined in blue have been fine-tuned from AlexNet \cite{krizhevsky2012} weights.}
\label{viralnet}
\vspace{-10pt}
\end{figure*}

The spatial transformer network $S$ is a multilayer convolutional network which produces an affine transformation as output. This affine transformation is applied to the input image and fed to the ranking network $R$. The ranking network takes a region-of-interest of the original image as input and produces a ranking score for the attribute in question. As a baseline, we implement the architecture described in \cite{singh2016end} and can be referred to for more details. This network is described in Figure \ref{viralnet}.\\

\begin{figure*}[t]
\centering
\includegraphics[width=1\textwidth]{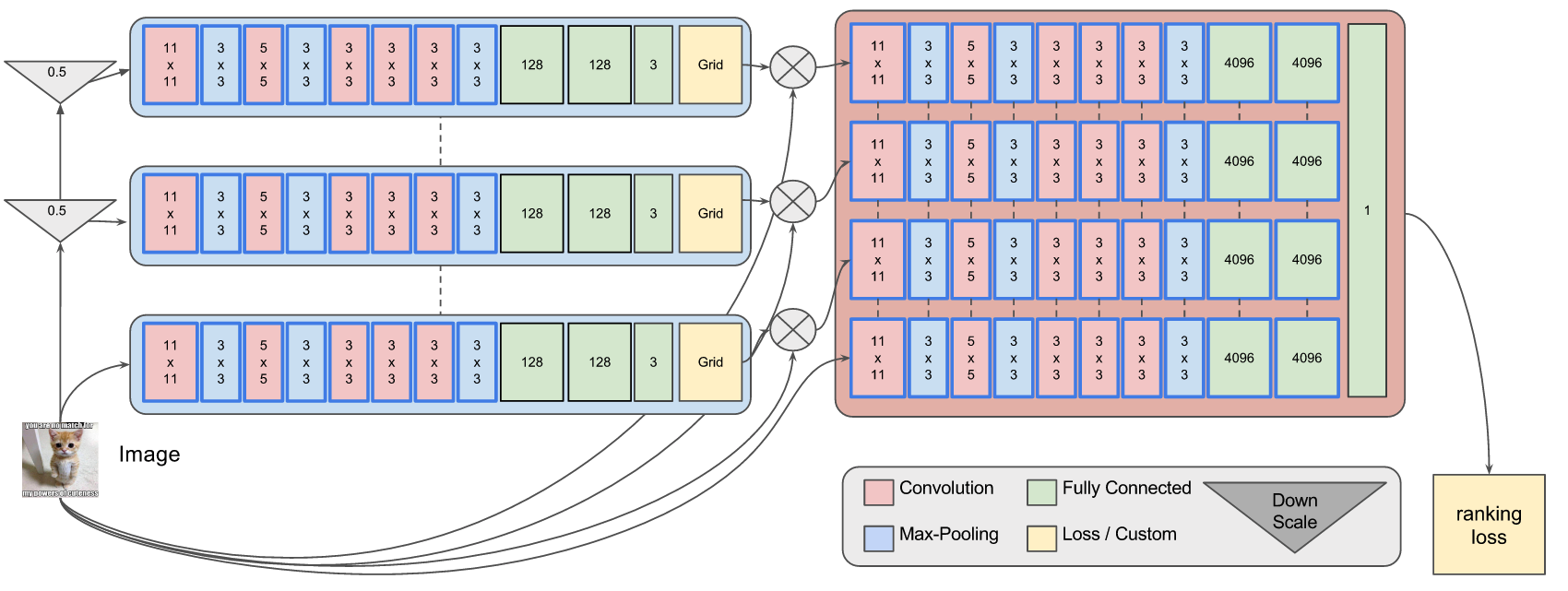}
\vspace{-10pt}
\caption{One half of the pyramidal multiscale \textbf{ViralNet-M} architecture. Here, instead of one spatial transformer network, we apply 3 spatial transformer networks at different scales and initialization settings for the image. The second half is ommitted due to space constraints but is identical in nature.}
\label{viralnetm}
\vspace{-10pt}
\end{figure*}

\textbf{Multiscale Region Pyramid:} A primary issue in applying pairwise STNs to imagery on the Internet is that cues in virality prediction are present at different scales, in contrast to a singular scale in regular attribute prediction. We can include response at multiple scales by adding filter banks with larger filter sizes to capture larger regions of context. However, this would increase the model complexity. Instead, we can downsample input images to simulate a larger detector. Hence, in addition to the first spatial transformation that operates on the entire image, we learn two additional spatial transformations, which operate at $0.5\times$ and $0.25\times$ downsampled version of the original image. Given an input image $\mathbf I$, we train 3 spatial transformer networks $S_1, S_2, S_3$ to provide input to the ranking network:
\begin{eqnarray}
S_1(\mathbf I) &=& \mathbf A_{\theta_1} \mathbf I_{\times 1}\\
S_2(\mathbf I) &=& \mathbf A_{\theta_2} \mathbf I_{\times 0.5} \\
S_3(\mathbf I) &=& \mathbf A_{\theta_3} \mathbf I_{\times 0.25}
\end{eqnarray}
Here, $\mathbf I_{x}$ represents image $\mathbf I$ downsampled by scale $x$. Contrary to \cite{singh2016end}, who provide image $\mathbf I$ and $S_1(\mathbf I)$ as both inputs to the ranker network, we provide 4 images - $\{\mathbf I, S_1(\mathbf I_{\times 1}), S_2(\mathbf I_{\times0.5}), S_3(\mathbf I_{\times0.25})\}$ to the feature extractor (see Figure \ref{viralnetm}). 

While the baseline network learns a ranking function $R(\mathbf I, S(\mathbf I))$, for each of the images in the input pair, we learn a ranking network $R(\mathbf I, S_1(\mathbf I_{\times1}), S_2(\mathbf I_{\times0.5}), S_3(\mathbf I_{\times0.25}))$ to generate a ranking score for each image in the input pair. To avoid an increase in number of parameters, the ranking function is formulated as a feature extractor $f_R$ followed by a ranker $r_R$. $f_R$ extracts a vector feature representation on each of the input sub-images, and concatenates them to get vector $t_R$ (see Equation \ref{concateqn}. $r_R$ then learns a linear ranking function on $t_R$ to obtain the final scalar output $v$ (see Figure \ref{viralnetm} for a better understanding of the same). This network is known as \textbf{ViralNet-M}, and is described in Figure \ref{viralnetm}.\\
\begin{eqnarray}
t_R(\mathbf I) &=& [ f_R(\mathbf I) , f_R(S_1(\mathbf I_{\times1})), f_R(S_2(\mathbf I_{\times0.5})), f_R(S_3(\mathbf I_{\times0.25})))]\\
v(\mathbf I) &=& \mathbf w_v^T t_R(\mathbf I) \label{concateqn}
\end{eqnarray}

\begin{figure*}[t]
\centering
\includegraphics[width=1\textwidth]{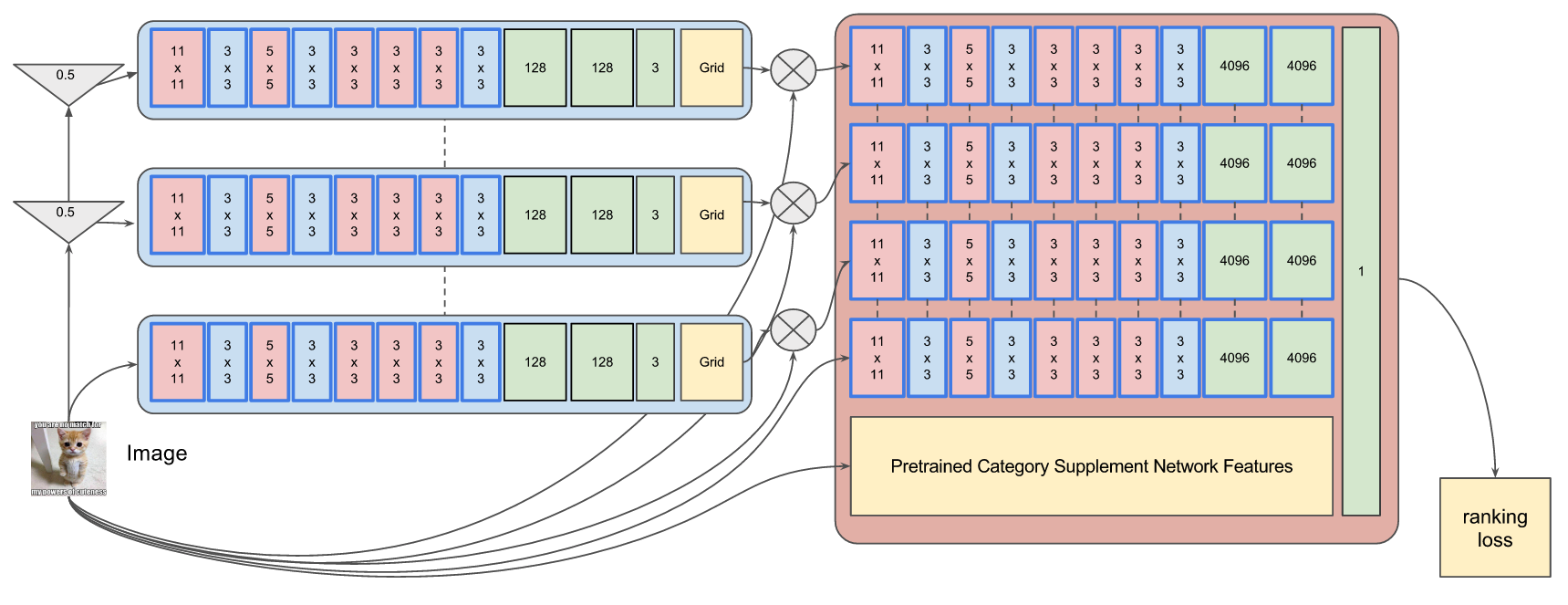}
\vspace{-10pt}
\caption{One half of the pyramidal multiscale \textbf{ViralNet-MC} architecture. This architecture retains all parameters from the \textbf{ViralNet-M} architecture except the final fully connected layer has an additional input of the category network. Similarly, for \textbf{ViralNet-C} architecture, the parameters are identical to the \textbf{ViralNet} architecture, but for the additional features from the category network as input to the final fully connected layer.}
\label{viralnetmt}
\vspace{-10pt}
\end{figure*}

\textbf{Category Supplement Network:} Apart from the label for the attribute strength, our target dataset contains labels for the category each submitted image belongs to. We utilize this additional supervision to improve prediction accuracy. For the training set, we train a neural network $C$ on images with category values as labels. We append copies of trained $C$ into our \textbf{ViralNet} and \textbf{ViralNet-M} architectures (one copy for each input image) and concatenate the features from $C$ into the final fully-connected layer to produce scalar values $v$. These architectures, named \textbf{ViralNet-C} and \textbf{ViralNet-MC} are described in Figure \ref{viralnetmt}.
\subsection{Datasets}
\label{datasets}
\hspace{5pt}\textbf{Viral Images Dataset:} As described in \cite{deza2015}, Lakkaraju et al. \cite{lakkaraju2013} crawled 132,000 entries submitted to Reddit over a period of four years, images of which form this dataset for virality prediction \cite{deza2015}. This dataset has 10,078 images, each with a relative score for virality. To create our train and validation sets, we sample images uniformly into training and test sets (80:20), and create image pairs within each of these. Our training data contains 10M image pairs, and validation data contains 1M image pairs, referred to as \textbf{Viral-Complete} in our results. For our test sets, we also use the two other datasets introduced by \cite{deza2015}, the `Viral and Non-Viral Images' dataset (referred to as \textbf{Viral-Pairs}) and `Random Pairs' dataset (referred to as \textbf{Viral-RandomPairs}), which contain image pairs distinct from the training/validation sets. For more details, please refer to \cite{deza2015} . To train the category dataset, we use the `Viral Categories Dataset' introduced by \cite{deza2015}.\\

\textbf{Image Popularity Dataset \cite{khosla2014}:} This dataset is the data utilized by Khosla et al. \cite{khosla2014} for their popularity analysis. It consists of 2.3M images sampled from Flickr and labeled as `popular' and `not-popular' according to their upvote measure. The three sub-categories for construction of the dataset are-\\

\hspace{5pt} \textit{One-per-user:} Images sampled from the Visual Sentiment Ontology dataset \cite{ontology} consisting of approximately 930K images from 400K users - collected by search Flickr for 3,244 adjective-noun-pairs. It represents the setting where different images correspond to different users, which is often the case in search results.\\

\hspace{5pt} \textit{User-mix:} This setting involves randomly selecting 100 users from the previous subset with between 10K and 20K public photos and accumulating all the corresponding images, resulting in approximately 1.4M images.\\

\hspace{5pt} \textit{User-specific:} The user-mix dataset is split into 100 different training and evaluation sets and results are averaged.\\

We use the same category model trained using the Virality data since there are no category labels present in this dataset.\\

\textbf{Attribute Learning Datasets:} As a comparative study, we also evaluate our model on the following relative attribute datasets-\\

\hspace{5pt} \textit{LFW-10 (LFW)}: This dataset contains images of 10 face attributes, with 1000 training images, 1000 test images and 500 pairs per attribute for both training and testing. We follow the splits mentioned in \cite{sandeep2014relative}. For category information we use the publicly available OpenFace model to generate features \cite{amos2016openface}.\\

\hspace{5pt} \textit{UT-Zap50K-1 (Zappos)}: This dataset contains 4 shoe attributes, with 50,025 shoe images, and 1388 training, 300 testing pairs per attribute. We follow the splits mentioned in \cite{yu2014fine}. We do not train category models on this dataset owing to the lack of category labels.\\

\hspace{5pt} \textit{OSR}: This dataset consists of 6 outdoor scene attributes with 2,688 total images. We follow the splits mentioned in \cite{yu2014fine,singh2016end}. For category information we use the publicly available Places205-AlexNet and Hybrid-CNN models \cite{zhou2014learning}.
\subsection{Training}
\subsubsection{Learning to Rank using Gradient Descent}
We train our \textbf{ViralNet} and \textbf{ViralNet-M} networks following the seminal work of RankNet \cite{burges2005learning}.\\ 

Specifically, the outputs $v(\mathbf I_1)$ and $v(\mathbf I_2)$ of the network are mapped to a probability $P$ via a logistic function $P = \frac{e^{v(\mathbf I_1)-v(\mathbf I_2)}}{1 + e^{v(\mathbf I_1)-v(\mathbf I_2)}}$ and then optimized using the cross entropy loss:
\begin{equation}
L_{rank}(\mathbf I_1,\mathbf I_2,y) = - y \cdot \log(P) - (1 - y) \cdot \log(1-P)
\end{equation}
Here, $y = 1$ if attribute strength is higher in $I_1$ compared to $I_2$ and vice-versa. For equality attributes, $y=0.5$. The nature of the loss function, as described in \cite{burges2005learning} is such that it asymptotes to a linear function and is more robust to noise compared to a quadratic function. 
\subsubsection{Spatial Transformer Networks for Ranking}
As mentioned in \cite{singh2016end}, it is possible that the spatial transformer networks predict regions outside the input image in order to minimize ranking error (by producing the same blank image input). To counter this, we also add the spatial transformation penalty as described by \cite{singh2016end}, which penalizes the spatial transformer networks if they produce regions outside the boundaries of the image. This spatial loss is given by - 
\begin{equation}
L_{spatial} = (C_x - s \cdot t_x)^2 + (C_y - s \cdot t_y)^2
\end{equation}
Here, $t_x$ and $t_y$ are the translations from the affine in horizontal and vertical directions, and $s$ is the isotropic scaling factor. $C_x$ and $C_y$ are the center coordinates of the input image. For \textbf{ViralNet} and \textbf{ViralNet-C}, the loss function is hence \cite{singh2016end}
\begin{equation}
L = (1 - \lambda_1)(1-\lambda_2) L_{rank} + \lambda_1 L^1_{spatial} + \lambda_2 L^2_{spatial}
\end{equation}
Here $\lambda_i$s are indicators which are 1 if the affine produced by either of the subnetworks is outside the boundaries of the image, and $L^i_{spatial}$ is the spatial loss incurred by the $i$th subnetwork. 
\subsubsection{Multiscale Pyramid Spatial Transformers}
For the multiscale pyramid networks, since there are 3 spatial transformer networks per subnetwork, the spatial transformer loss is calculated for each of these networks, and the total loss for networks \textbf{ViralNet-M} and \textbf{ViralNet-MC} is given by
\begin{multline}
L_M  =  (\prod_{i=1}^3 (1 - \lambda_{1,i})) \cdot (\prod_{i=1}^3 (1-\lambda_{2,i})) L_{rank} \\
 + \sum_{i=1}^3 \lambda_{1,i} \cdot L^{1,i}_{spatial} + \sum_{i=1}^3 \lambda_{2,i} L^{2,i}_{spatial} 
\end{multline}
Here, $\lambda_{i,j}$ denotes if the affine produced by the $j$th spatial network in subnetwork $i$ is out of image boundaries, and $L^{i,j}_{spatial}$ is the spatial loss produced by the $j$th spatial network in subnetwork $i$. During training time, each input image is cropped with 90\% area and different crops are fed to different scales, to prevent overfitting and to map to different transformations.\\

We train these networks using backpropagation on the relative attribute training data, and the category networks are trained beforehand either from publicly available models or using the data described in the earlier section. The category models are not fine-tuned during the attribute training process, their weights are kept fixed. Additionally, the spatial loss $L_{spatial}$ applies only to the spatial transformer networks and not the ranker networks. All initialization schemes are following \cite{singh2016end}.
\begin{table*}[h]
\centering
\small
\begin{tabular}{|l|c|c|c|}
\hline
\multirow{2}{*}{\textbf{Algorithm}}            & \multicolumn{3}{c|}{{\textbf{Percentage Accuracy}}} \\ \cline{2-4}
& Viral-Complete & Viral-Pairs & Viral-RandomPairs \\ \hline
Chance               & \multicolumn{3}{c|}{50}     \\ \hline
SVM + Image Features \cite{deza2015} & 53.40 & 61.60 & 58.49 \\ \hline
RankSVM + 2x2 Dense HOG Features & 52.75 & 58.81 & 56.92 \\ \hline
RankSVM + AlexNet fc7 Features \cite{krizhevsky2012} & 54.41 & 61.93 & 58.58 \\ 
RankSVM + VGGNet-16 fc7 Features \cite{simonyan2014} & 55.18 & 63.01 & 59.15 \\ \hline
Human \cite{deza2015} & - & 71.76 & 60.12 \\ 
Popularity API \cite{khosla2014} & - & - & 51.12 \\
Human Annotated Atts. -5 \cite{deza2015} & - & - & 65.18 \\
SVM + Deep Attributes-5 \cite{deza2015} & - & - & 68.10 \\ \hline
Xiao and Lee \cite{xiao2015discovering} & 63.25 & 75.23 & 73.22 \\ \hline
\textbf{ViralNet} \cite{singh2016end} & 65.87 & 76.20 & 74.38 \\
\textbf{ViralNet-M} & 66.15 & 77.01 & 75.52 \\
\textbf{ViralNet-C} & \textbf{67.88} & \textbf{77.60} & \textbf{76.78} \\
\textbf{ViralNet-MC} & \textbf{68.09} & \textbf{78.38} & \textbf{76.95} \\ \hline 
\end{tabular}
\caption{Table summarizing our empirical results on the Viral Images dataset. The results on different baselines have been replicated from the referred papers in each entry. We observe that on this dataset we comfortably outperform the existing state-of-the-art by a relative improvement of 12\% on the toughest split, and relative improvements of \textbf{24\%} and \textbf{23.3\%} on the easier splits.}
\label{res:virality}
\vspace{-20pt}
\end{table*}
\section{Results}
Our experiments were carried out using a combined architecture of TensorFlow \cite{tensorflow} written in Python and Caffe \cite{caffe} with the Python framework. All experiments were run on NVIDIA Tesla k40 machines. All mentioned model weights were obtained from the Caffe Model-Zoo \cite{modelzoo}. We follow the learning rate, initialization and implementation details are identical to \cite{singh2016end} for the single-scale networks.

For the multi-scale pyramids, the scale values were initialized to 1, 0.5 and 0.25 times the initialization for single-scale network scale values respectively.\\

\textbf{Baselines:} Our baseline technique \textbf{ViralNet} when applied to the relative attribute datasets, comprises the exact experiments reported by \cite{singh2016end}. Additionally, we compare with the work of Xiao and Lee \cite{xiao2015discovering} in discovering spatial extents for attributes, the baselines reported by Deza and Parikh \cite{deza2015} and features extracted from popular image classification architectures such as AlexNet \cite{krizhevsky2012} and VGGNet \cite{simonyan2014} as well as shallow image features plugged into a RankSVM \cite{joachims2002}.
\subsection{Viral Images Dataset}
Our experiments begin with the Viral Images Dataset \cite{deza2015}, which is a comprehensive inference dataset for virality prediction. We find that even the baseline network \cite{singh2016end} outperforms the existing state-of-the-art on this dataset \cite{deza2015} comfortably, and our final model \textbf{ViralNet-MC} gives a significant improvement over the baseline. Specifically, we observe that the addition of category features from the network provide positive signal to the base networks. We visualize the region networks for our multiscale and baseline networks in more details in the next section. The results on this dataset are described in more detail in Table \ref{res:virality}.
\subsection{Popular Images Dataset}
The next experiments we perform are on the Popular Images Dataset \cite{khosla2014}. On this dataset as well, we outperform the existing state-of-the-art comfortably on all splits. However, we find that the category subnetwork, which is trained on the `Viral Categories' dataset, provides less supervision compared to the previous dataset. Our complete results are reported in Table \ref{res:popularity}.
\begin{table*}[h]
\centering
\small
\begin{tabular}{|l|c|c|c|}
\hline
\multirow{2}{*}{\textbf{Algorithm}}            & \multicolumn{3}{c|}{{\textbf{Percentage Accuracy}}} \\ \cline{2-4}
                    & One-per-user    & User-mix    & User-specific \\ \hline
GIST \cite{khosla2014}          & 50.32       & 56.61     & 57.74     \\ 
Color Histogram \cite{khosla2014}     & 56.13       & 57.58     & 59.07     \\
Texture \cite{khosla2014}         & 59.91       & 61.01     & 63.02     \\
Color Patches \cite{khosla2014}     & 60.22       & 61.09     & 64.19     \\
Gradient \cite{khosla2014}        & 62.12       & 63.04     & 64.47     \\
DeCAF \cite{decaf} Features \cite{khosla2014} & 63.89     & 65.05     & 64.89     \\
Objects \cite{khosla2014}         & 62.01       & 65.17     & 65.32     \\
Combined \cite{khosla2014}        & 64.02       & 65.98     & 66.18     \\ \hline
RankSVM + AlexNet fc6 Features \cite{krizhevsky2012} & 66.01 & 67.75    & 67.02     \\ 
RankSVM + VGGNet-16 fc6 Features \cite{simonyan2014} & 67.15 & 68.97    & 67.28     \\ \hline
Xiao and Lee \cite{xiao2015discovering} & 70.02       & 71.32     & 72.86     \\ \hline
\textbf{ViralNet} \cite{singh2016end}   & 71.89       & 74.08     & 75.33     \\
\textbf{ViralNet-M}           & 72.57       & 75.59     & 76.99     \\
\textbf{ViralNet-C}           & 72.84       & 75.52     & 76.97     \\
\textbf{ViralNet-MC}          & \textbf{73.01}  & \textbf{76.07}& \textbf{77.78}\\ \hline 
\end{tabular}
\caption{Table summarizing our empirical results on the Popular Images dataset. The results on different baselines have been replicated from the referred papers in each entry. We observe that on this dataset we comfortably outperform the existing state-of-the-art by a relative improvement of 8\%.}
\label{res:popularity}
\vspace{-20pt}
\end{table*}
\subsection{Relative Attribute Datasets}
We also evaluate our algorithm on relative attribute datasets to observe the performance benefits of category and multiscale interventions in end-to-end spatial transformer networks. As expected, we observe that the category supervision provides positive signal to the attribute prediction. Additionally, the multiscale intervention also provides additional signal in datasets which have multiple scales of localization, such as \textbf{OSR} and \textbf{LFW-10}. We report the mean prediction accuracy over all additional relative attribute datasets in Table \ref{res:attributes}. Note here that the architecture for \textbf{ViralNet} is identical to the architecture introduced by \cite{singh2016end}.
\begin{table}[h]
\centering
\small
\begin{tabular}{|l|c|c|c|}
\hline
\multirow{2}{*}{\textbf{Algorithm}}            & \multicolumn{3}{c|}{{\textbf{Mean Percentage Accuracy}}} \\ \cline{2-4}
& \textbf{LFW} & \textbf{Zappos} & \textbf{OSR} \\ \hline
Parikh and Grauman \cite{parikh2011} + CNN \cite{singh2016end} & 74.61 & 94.69 & 94.98 \\ \hline
Xiao and Lee \cite{xiao2015discovering} & 84.66 & 95.47 & 92.16 \\ \hline
End-to-End Localization \cite{singh2016end} & 86.91 & 95.78 & 97.02 \\ \hline
\textbf{ViralNet-M} & 86.94 & \textbf{96.01} & 96.68 \\
\textbf{ViralNet-C} & 87.15 & - & 97.02 \\
\textbf{ViralNet-MC} & \textbf{87.67} & - & \textbf{97.55} \\ \hline 
\end{tabular}
\caption{Table summarizing our empirical results on the several relative attribute learning datasets. The results on different baselines have been replicated from the referred papers in each entry. We observe that the category supplement networks provide positive signal in prediction, increasing accuracy across datasets.}
\label{res:attributes}
\vspace{-20pt}
\end{table}
\subsection{Qualitative Experiments}
\begin{figure*}[t]
\centering
\includegraphics[width=0.8\textwidth]{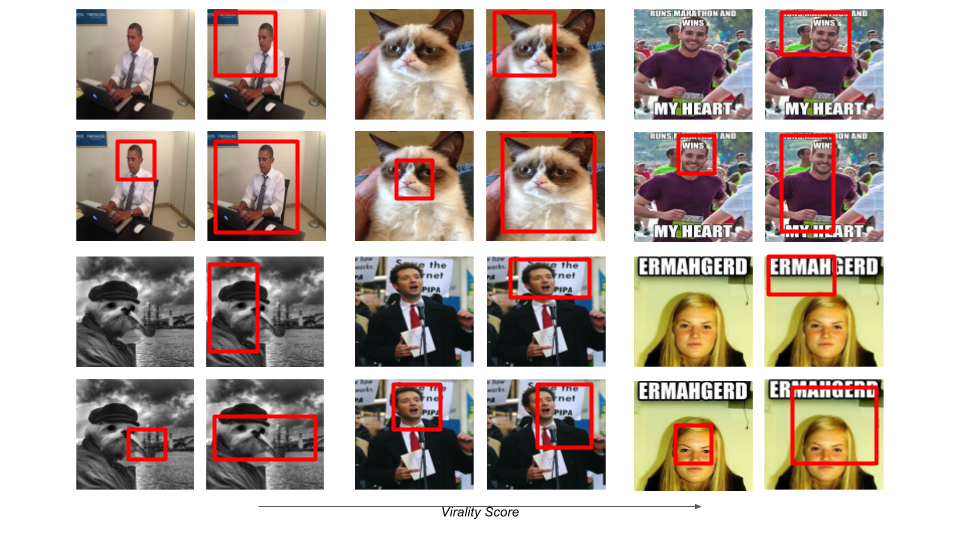}
\caption{We examine the qualitative effectiveness of the spatial transformer networks in this experiment. For our \textbf{ViralNet-MC} model, we feed images of varying degrees of virality metric and observe the regions fed into the ranking networks at different scales. We find that at different scales, these regions focus on different aspects of input images. For each image group, the top-left is the original image, bottom-left is the localization at 0.25 scale, top-right is the localization at 0.5 scale and bottom-right is the localization at original scale. We see that the model latches on to faces at smaller scales, and text if present.}
\label{fig:localization}
\end{figure*}

To further understand the qualitative nature of our results and interpret the decision surfaces produced by our models, we perform several visualization experiments. Singh et al \cite{singh2016end} provide a qualitative localization experiment by visualizing the region of interest fed to the ranker networks from the spatial transformer networks, for both the single-scale and multi-scale pyramidal networks. We observe that our scale initialization is critical for the networks to feed different regions of interests at various scales in these networks. The visualizations in the virality and popularity datasets provide significant insights into the regions the convolutional networks are focusing on during prediction. We find that the networks latch on to areas of interest which can be visualized as important to virality - the presence of text, bright colors and humans in images promote virality, and the localization areas returned by these networks agree with these propositions. Figure \ref{fig:localization} provides sample visualizations.\\

\begin{figure}[h]
\centering
\includegraphics[width=0.8\linewidth]{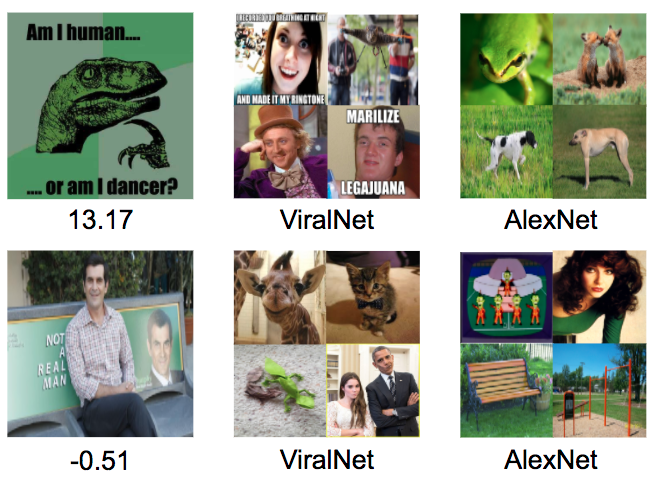}
\caption{Nearest Neighbours for two sample inputs to both images in subspace of pre-final activations with their representative virality scores. We see that our architecture ViralNet localizes more closely to the viral categories, whereas AlexNet localizes more closely to visual representations and objects.}
\label{fig:neighbors}
\end{figure}

To understand in detail the contribution of the category features in attribute prediction, we also perform another qualitative visualization experiment. We visualize the nearest neighbors in the penultimate layers of both the \textbf{ViralNet} and \textbf{AlexNet} models for different values of virality strengths. In images classified by either networks, we examine the nearest neighbors in the training set for each, and observe that the  improvement is achieved because of localization in the space of abstract attribute information. See Figure \ref{fig:neighbors} for sample visualizations.
\subsection{Ablation Studies}
Our proposed model takes focuses on the usage of spatial transformer networks to localize regions important for the prediction of virality and popularity as attributes. Additionally, our formulation of tackling the prediction task as a relative attribute learning task focuses on the relative nature of attribute values as a stronger signal for modeling virality. In order to justify our two interventions in this problem, we conduct ablation studies for both interventions, first by removing the spatial transformers completely and using just a Siamese convolutional network (similar to \cite{chopra2005}) with the RankNet loss \cite{burges2005learning}, initalized with the AlexNet \cite{krizhevsky2012} and VGGNet-16 \cite{simonyan2014} model weights. Secondly, we also compare the straightforward formulation of training a regressor on existing architectures AlexNet and VGGNet-16 on the virality values. We summarize the results of this ablation experiment in Table \ref{res:ablation}, and our results justify both the choices as we see that the ablation models perform significantly poorly compared to the \textbf{ViralNet} architectures.
\begin{table}[h]
\centering
\small
\begin{tabular}{|l|c|c|c|}
\hline
\multirow{2}{*}{\textbf{Algorithm}}            & \multicolumn{3}{c|}{{\textbf{Percentage Accuracy}}} \\ \cline{2-4}
& Complete & Viral-Pairs & Random-Pairs \\ \hline
AlexNet \cite{krizhevsky2012} Regression & 53.01 & 57.72 & 54.19 \\ 
VGGNet-16 \cite{simonyan2014} Regression & 56.12 & 60.11 & 58.23 \\ \hline
AlexNet \cite{krizhevsky2012} Siamese & 58.17 & 61.11 & 59.87 \\ 
VGGNet-16 \cite{simonyan2014} Siamese & 60.23 & 66.15 & 63.15 \\ \hline
\textbf{ViralNet} & \textbf{65.87} & \textbf{76.20} & \textbf{74.38} \\ \hline
\end{tabular}
\caption{Table summarizing the results of our ablation studies on the Viral Images dataset. We find that both our interventions to the existing models provide significant improvements in virality prediction.}
\label{res:ablation}
\vspace{-20pt}
\end{table} 
\section{Conclusion}
In this paper, we presented a novel interpretation of the problem of predicting virality as a pairwise attribute learning task, and presented a novel multiscale pyramidal deep architecture to predict virality with a significant improvement (19.1\% average relative increment) over the existing state-of-the-art. Additionally, our network is able to localize several different areas of interest for each image which provides significant insights into the qualitative understanding of virality, which are vital in the construction and easy caching of viral content on the Internet.\\

The augmentation with category information indicates the value of supervision via well-defined labels on the performance of an amorphous recognition task, and is suggestive of similar improvements on other abstract learning tasks such as memorability prediction and urban perception. It has long been understood that the number of shares and connectivity of a content outlet has a significant impact on a shared image's popularity, and our work provides evidence that there is significant contribution of the content as well in its popularity.\\

In conclusion, we provide a qualitative and quantitative improvement over the existing work on image virality prediction and abstract attribute learning, that can be generalizable to any abstract classification task. We believe this technique of visualizing network regions will be useful in understanding relative attributes and classification, as well as further our understanding what promotes virality on the Internet.
{\small
\bibliographystyle{ACM-Reference-Format}
\balance
\bibliography{virality} }

\end{document}